 \documentclass[pmlr,twocolumn,10pt]{jmlr} %

\usepackage{booktabs}
\usepackage{siunitx}
\usepackage{adjustbox}

\newcommand{\equal}[1]{{\hypersetup{linkcolor=black}\thanks{#1}}}

\usepackage[textsize=tiny]{todonotes}
\theorembodyfont{\upshape}
\theoremheaderfont{\scshape}
\theorempostheader{:}
\theoremsep{\newline}

\jmlrvolume{225}
\jmlryear{2023}
\jmlrsubmitted{LEAVE UNSET}
\jmlrpublished{LEAVE UNSET}
\jmlrworkshop{Machine Learning for Health (ML4H) 2023} %

\title[MoCo-Transfer]{MoCo-Transfer: 
Investigating out-of-distribution contrastive learning for limited-data domains}

\author{%
\Name{Yuwen Chen}\equal{These authors contributed equally} \Email{yuwenc2@andrew.cmu.edu}\\
\addr Carnegie Mellon University, USA
\AND
\Name{Helen Zhou}\footnotemark[1] \Email{hlzhou@cmu.edu}\\
\addr Carnegie Mellon University, USA
\AND
\Name{Zachary C. Lipton} \Email{zlipton@cmu.edu}\\
\addr Carnegie Mellon University, USA
}

\begin{document}

\maketitle
\begin{abstract}

Medical imaging data is often siloed within hospitals, limiting the amount of data available for specialized model development. With limited in-domain data, one might hope to leverage larger datasets from related domains. In this paper, we analyze the benefit of transferring self-supervised contrastive representations from moment contrast (MoCo) pretraining on out-of-distribution data to settings with limited data. We consider two X-ray datasets which image different parts of the body, and compare transferring from each other to transferring from ImageNet. We find that depending on quantity of labeled and unlabeled data, contrastive pretraining on larger out-of-distribution datasets can perform nearly as well or better than MoCo pretraining in-domain, and pretraining on related domains leads to higher performance than if one were to use the ImageNet pretrained weights. Finally, we provide a preliminary way of quantifying similarity between datasets.

\end{abstract}

\vspace{-1em}
\section{Introduction} \label{sec:intro}

Research efforts in medical imaging are often contingent upon the availability and quantity of pertinent data.
For example, the public release of large datasets containing chest radiographs \citep{wang2017chestx,irvin2019chexpert,johnson2019mimic}, whole-slide pathology images \citep{bejnordi2017diagnostic,bandi2018detection}, and brain MRIs \citep{petersen2010alzheimer,van2013wu,rsna-intracranial-hemorrhage-detection}  has catalyzed computer vision research in those domains. However, many specialized imaging domains continue to operate with limited data reservoirs, thereby receiving comparatively scant attention from the machine learning research community.
Hospital data is often siloed, expert labeling is expensive, and researchers may only have access to a few pockets of data. Moreover, variations in data collection policies, equipment, and imaging modalities can lead to disparate data distributions in the same specialty over time and across hospitals \citep{doi:10.1148/radiol.2020192224,gupta2021addressing,zhou2023evaluating}. 

In this work, we study the extent to which contrastive pretraining on unlabeled out-of-domain medical images can learn representations that improve performance when there is limited labeled and unlabeled data. In particular, we examine whether images in related domains may help in-domain. To summarize our contributions, we:
\begin{itemize}
    \vspace{-0.5em}
    \item Show that self-supervised contrastive learning on unlabeled data from related domains can improve in-domain performance when limited in-domain data is available.
    \vspace{-0.5em}
    \item Characterize the impact of the amount of labeled and unlabeled data on the marginal benefit of using different MoCo pretrained representations.
    \item Find that freezing all but the last layer of weights can improve performance when finetuning on small amounts of labeled data.
    \vspace{-0.5em}

    \item Provide a preliminary technique for quantifying similarity between datasets.

\end{itemize}

\vspace{-1em}
\section{Related Work}\label{sec:related_work}

Common strategies for learning from limited imaging data include using out-of-domain data for transfer learning, creating synthetic samples to boost sample size for data augmentation, and leveraging unlabeled data for semi-supervised learning.

\paragraph{Transfer Learning on Medical Images}
Modern computer vision models have been developed on enormous 
datasets 
containing millions of natural images
\citep{deng2009imagenet,szegedy2015going,he2016deep,huang2017densely}.
Although natural images differ from medical images in tasks of interest, granularity, variability, and other image statistics, empirical works have noted improved performance from initializing models with ImageNet pretrained weights before finetuning on medical imaging datasets
\citep{Xie_2018_ECCV_Workshops,app10134523,10.1145/3450439.3451867}. 
On large retinal fundus and chest X-ray imaging datasets (hundreds of thousands of images), detailed analyses by 
\citet{NEURIPS2019_eb1e7832} showed that ImageNet pretraining does not necessarily improve performance, but yields different representations than random initialization and 
can improve training speed and convergence.

Later works have also studied how well pretrained representations transfer between different medical imaging domains. 
\citet{WEN2021103145} studied the benefit of transferring supervised pretraining among classification and segmentation tasks on X-ray, fundoscopic, CT, and MRI datasets, finding that ImageNet pretraining was still the best option. \citet{butoi2023universeg} trained a UniverSeg segmentation model on 53 datasets encompassing 26 medical domains and 16 imaging modalities, finding that the segmentation capabilities extend to previously unseen domains.

\paragraph{Data Augmentation} 
Affine transformations, intensity manipulations, rotating, blurring, and cropping are often used to augment training data \citep{10.1117/12.2293971,8363576, chlap, GARCEA2023106391}. These techniques broaden the range of images to which models are exposed, and encourage robustness to such transformations. Various works have also utilized mixup \citep{zhang2017mixup,eaton2018improving} and generative models such as GANs \citep{goodfellow2020generative,FRIDADAR2018321,10.1371/journal.pone.0267976} and stable diffusion \citep{rombach2022high,trabucco2023effective} in order to augment the training data.

\paragraph{Contrastive Learning} For settings with limited labeled data but substantial unlabeled data, contrastive self-supervised methods such as SimCLR \citep{chen2020big,chen2020simple}, PIRL \citep{misra2020self}, and MoCo \citep{he2020momentum,chen2020improved} have been used to pretrain in a task-agnostic self-supervised manner before finetuning with the supervised objective of interest \citep{krishnan2022self}. Contrastive learning has been adapted for multiple-instance learning \citep{Azizi_2021_ICCV}, where there are multiple images of the underlying pathology per patient case, as well as
volumetric medical images \citep{NEURIPS2020_949686ec}, 
where structural similarity across volumes is leveraged to learn representations of local regions useful for segmentation. \citet{sowrirajan2021mococxr} explore the use of MoCo pretraining on chest X-rays for classification of chest X-rays, finding some performance benefit in settings with limited labeled data. \citet{anton2022} find that ImageNet pretrained self-supervised models generalize better than supervised pretraining, and across chest X-ray, retinal fundus, and breast histology datasets, find that pretrained models improve in-domain performance but deteriorate out-of-domain performance.

\begin{figure*}[ht]
\centering
\includegraphics[width=2\columnwidth]{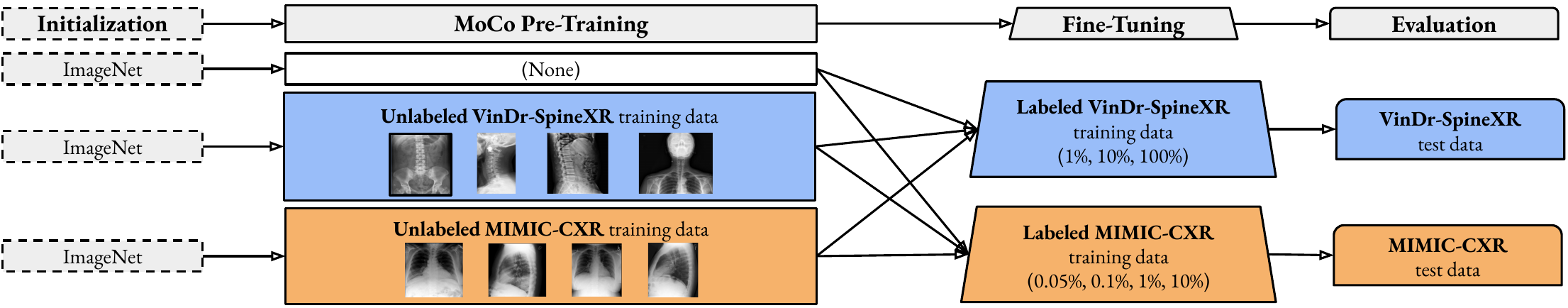}
\vspace{-1em}
\caption{Datasets used in each stage of the MoCo-Transfer training and evaluation pipeline, from initial weights, to self-supervised contrastive pretraining, to fine-tuning, to evaluation.}
\vspace{-2em}
\label{fig:moco_transfer_pipeline}
\end{figure*}

\vspace{-1em}
\section{Data}\label{sec:data}

We analyze two medical imaging domains: spine X-rays and chest X-rays. In contrast to prior work studying out-of-distribution transfer \citep{NEURIPS2019_eb1e7832,anton2022}, these domains have more qualitative similarities, as they are related by imaging modality (X-ray) and have overlapping imaging regions in the body (see Figure \ref{fig:moco_transfer_pipeline} for example images).

\paragraph{VinDr-SpineXR}

The VinDr-SpineXR 
dataset contains 10,468 spine X-ray images annotated by experienced radiologists \citep{nguyen2021vindr}. The task of interest is binary classification of whether an abnormality is present. We use the default test set (20\% of the data), and split the default training set 70-30 to create train and validation splits. 

\paragraph{MIMIC-CXR}
The MIMIC Chest X-ray JPG Database is much larger, containing 377,110 chest radiographs \citep{johnson2019mimiccxrjpg} with 14 diagnostic labels extracted from corresponding radiology reports. We randomly split the data into 60-20-20 for train, validation, and test sets.

\vspace{-1em}
\section{Methods}\label{sec:methods}

\paragraph{Model Architecture} 
Like previous works \citep{nguyen2021vindr,johnson2019mimiccxrjpg,seyyed2020chexclusion,sowrirajan2021mococxr}, we use DenseNet-121 \citep{huang2017densely}. See Appendix \ref{app:training_details} for hyperparameter details.
 
\paragraph{Training Objectives and Evaluation}
Data augmentation transformations, same as \citet{chen2020improved} and \citet{sowrirajan2021mococxr}, are used for MoCo pretraining. An augmented version of the same image is a positive pair, and a different image is a negative pair.
MoCo pretraining seeks to minimize the InfoNCE loss $L_{q} = -\log \frac{\exp(qk_{+} / \tau)}{\sum_{i=0}^K \exp(qk_i / \tau)}$, which encourages the output representation $q$ of an input image to be similar to its positive key $k_{+}$ and dissimilar to other keys in the dictionary, for some temperature hyper-parameter $\tau$ and number of samples $K$. MoCo is used instead of other methods such as SimCLR \citep{chen2020improved} for computational reasons, as it allows for smaller batch sizes. Binary cross entropy loss is used in both datasets. Since MIMIC-CXR involves multi-label classification, we evaluate the model by calculating weighted AUROC, weighted by number of samples per class.

\paragraph{Experiment Setup}
 
Each experiment initializes with pretrained ImageNet weights for DenseNet-121, MoCo pretrains on a dataset, and finetunes and evaluates on the dataset of interest. 
(Figure \ref{fig:moco_transfer_pipeline}). MoCo pretraining is done on (1) VinDR-SpineXR, (2) MIMIC-CXR, and (3) nothing. finetuning is done on VinDR-SpinXR and MIMIC-CXR either (a) end-to-end, where all weights can be tuned, or (b) linearly, where all but the last layer of weights are frozen. End-to-end finetuning gives greater flexibility to adapt to the downstream task, and linear finetuning preserves more of the MoCo representation.

To investigate the benefit of MoCo pretraining under different amounts of labeled data, we finetune with 1\%, 10\%, and 100\% of the VinDR-SpineXR training data
(58 to 5,872 samples), 
and 0.05\%, 0.1\%, 1\%, and 10\% of the MIMIC-CXR training data
(112 to 22,572 samples). 
For small percentages in linear finetuning, we also experiment with MoCo pretraining using as many unlabeled data points as labeled data points in order to understand the relative impact of unlabeled data quantity.
Three downsamplings are done for each percentage of labeled data, and confidence intervals around the average test performance
are computed by bootstrap resampling the test set 500 times, and taking the 5\% and 95\% percentiles.

\paragraph{Quantifying Dataset Similarity}
While prior work has shown that pretraining on chest X-rays deteriorates out-of-domain performance on retinal fundus and breast histology images \citep{anton2022}, we argue that there is more nuance to consider when selecting domains to transfer between. Although the notion of a ``related'' imaging domain can be somewhat nebulous, we propose a preliminary method for quantifying how close two datasets might be. 

The VinDR-SpineXR, MIMIC-CXR, and ImageNet datasets are compared pairwise by performing Singular Vector Canonical Correlation Analysis (SVCCA) on the activations of 2,000 randomly sampled data points from each dataset. SVCCA has been used to study latent representations across different models \citep{raghu2017svcca}. It computes the similarity the neural activations of a given layer in two different models by (1) using singular value decomposition to get the most important directions of the original subspaces, and (2) linearly transforming these subspaces to be aligned as possible and computing correlation coefficients. In addition to the typical usage of SVCCA for comparing models, we hypothesize that SVCCA could be used to compare datasets, by comparing the distribution of activations resulting from inputs from different datasets.

\vspace{-1em}
\section{Results \& Discussion}\label{sec:results}

\begin{figure*}[ht]
    \centering
    \includegraphics[width=2\columnwidth]{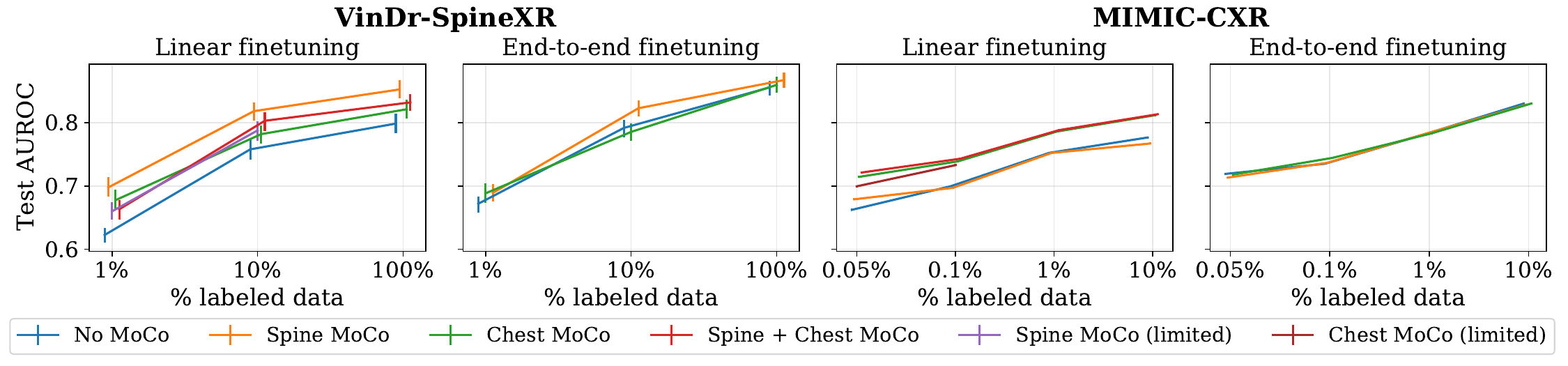}
    \vspace{-1.4em}
    \caption{Test AUROC vs. \% of labeled data used for linear and end-to-end finetuning on VinDr-SpineXR (Spine) and MIMIC-CXR (Chest). MoCo (limited) refers to MoCo pretraining on the same subset of data that is labeled (for finetuning). Error bars give 95\% confidence intervals based on bootstrapping with 500 resamplings, and averages are computed across three seeds.}

    \vspace{-2em}
    \label{fig:testauroc}
\end{figure*}

\paragraph{Linear finetuning} In linear finetuning on both MIMIC-CXR (Chest) and VinDR-SpineXR (Spine), the in-domain MoCo pretrained representations outperform the out-of-distribution datasets. Among the out-of-distribution datasets, however, pretraining on related X-ray datasets outperforms pretraining on ImageNet weights (Appendix Tables \ref{tab:vindr_table} and Tables \ref{tab:mimic_table}).

Additionally, in VinDr-SpineXR, if one limits the amount of unlabeled in-domain data for MoCo to match the size of the labeled dataset, using a larger quantity of out-of-distribution unlabeled MIMIC-CXR data for MoCo can even slightly improve over using in-domain data for pretraining (Figure \ref{fig:testauroc}). These improvements over ImageNet and even in-domain data may be due to similar structures in the Chest and Spine X-rays (e.g. vertebrae and ribs). MoCo pretraining on Spine + Chest data results in similar performance as in-domain Chest MoCo on both datasets. Since there is substantially more Chest data than Spine data, the learned representations may be biased heavily towards Chest data.

\paragraph{End-to-end}
When finetuning the model end-to-end, the benefits of MoCo pretraining versus ImageNet initialization become less apparent. Since the MoCo representations are no longer strictly preserved, end-to-end finetuning may be fitting more closely to the labeled data, and with enough iterations, may forget the initial MoCo representations. We also note that the number of epochs in end-to-end training can greatly affect the relative performance of different initializations (Figures \ref{fig:vindr_vs_epoch} and \ref{fig:mimiccxr_vs_epoch}), motivating our hyperparameter selection over epochs.

\paragraph{Linear vs. End-to-end} As expected, in both finetuning strategies, more labeled data improves the overall test AUROC (Figure \ref{fig:testauroc}). Interestingly, in settings with limited labeled data (e.g. VinDr-SpineXR with 1\% and 10\% labeled and MIMIC with 1\% labeled), linear finetuning often outperforms end-to-end finetuning (Appendix Tables \ref{tab:vindr_table} and \ref{tab:mimic_table}).

\paragraph{Dataset similarity}
\begin{figure}[t]
    \centering
    \includegraphics[width=0.9\columnwidth]{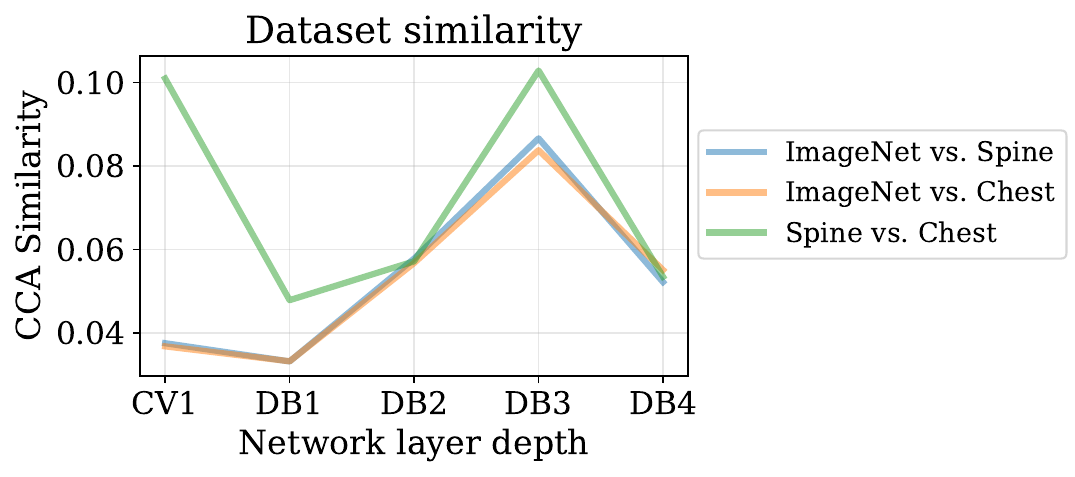}
    \vspace{-1.5em}
    \caption{Pairwise CCA similarity between datasets across different layers in DenseNet-121. ``CV1'' = first convolution layer, and ``DB$x$'' = last activation layer in the $x$th dense block.}
    \vspace{-2em}
    \label{fig:dataset_cca}
\end{figure}

As shown in Figure \ref{fig:dataset_cca}, VinDR-CXR and MIMIC-CXR obtain higher a CCA similarity score than ImageNet with either of the X-ray datasets, across several layers of the DenseNet-121 architecture. This is consistent with Chest MoCo having intermediate performance between ImageNet pretraining and Spine MoCo pretraining when evaluated on the Spine dataset.

\vspace{-1em}
\section{Conclusion}\label{sec:discussion}

We investigate the transferability of MoCo representations from out-of-distribution datasets in settings with limited data. Representations learned from related domains, such as chest X-rays and spine X-rays, may benefit in-domain performance when in-domain data is scarce, and improve over standard pretrained weights on ImageNet. This suggests that some notion of relatedness may be helpful in determining which out-of-distribution imaging datasets might be promising to build upon, and we provide a preliminary technique for quantifying relatedness, consistent with our observations of which datasets did observe benefits from transfer learning. In settings where there is substantial in-domain unlabeled data but limited labeled data, we find that linear finetuning can often yield better performance than end-to-end finetuning. In future work, we would like to explore a broader range of imaging modalities and targets.

\acks{
We gratefully acknowledge the NSF (FAI 2040929 and
IIS2211955) Amazon AI, UPMC, Highmark Health,
Abridge, Ford, Mozilla, the PwC Center, the Block Center, the Center for Machine Learning and Health, and the
CMU Software Engineering Institute (SEI) via Department
of Defense contract FA8702-15-D-0002, for their generous
support of ACMI Lab’s research. This research was also supported in part by the Paul and Daisy Soros Fellowship and the National Science Foundation Graduate
Research Fellowship Program under grant numbers DGE1745016 and DGE2140739.
}

\bibliography{zhou23}

\begin{thebibliography}{46}
\providecommand{\natexlab}[1]{#1}
\providecommand{\url}[1]{\texttt{#1}}
\expandafter\ifx\csname urlstyle\endcsname\relax
  \providecommand{\doi}[1]{doi: #1}\else
  \providecommand{\doi}{doi: \begingroup \urlstyle{rm}\Url}\fi

\bibitem[Alzubaidi et~al.(2020)Alzubaidi, Fadhel, Al-Shamma, Zhang, Santamaría, Duan, and R.~Oleiwi]{app10134523}
Laith Alzubaidi, Mohammed~A. Fadhel, Omran Al-Shamma, Jinglan Zhang, J.~Santamaría, Ye~Duan, and Sameer R.~Oleiwi.
\newblock Towards a better understanding of transfer learning for medical imaging: A case study.
\newblock \emph{Applied Sciences}, 10\penalty0 (13), 2020.
\newblock ISSN 2076-3417.
\newblock \doi{10.3390/app10134523}.
\newblock URL \url{https://www.mdpi.com/2076-3417/10/13/4523}.

\bibitem[Anton et~al.(2022)Anton, Castelli, Chan, Outters, Tang, Cheung, Shukla, Walambe, and Kotecha]{anton2022}
Jonah Anton, Liam Castelli, Mun~Fai Chan, Mathilde Outters, Wan~Hee Tang, Venus Cheung, Pancham Shukla, Rahee Walambe, and Ketan Kotecha.
\newblock How well do self-supervised models transfer to medical imaging?
\newblock \emph{Journal of Imaging}, 8\penalty0 (12):\penalty0 320, Dec 2022.
\newblock ISSN 2313-433X.
\newblock \doi{10.3390/jimaging8120320}.
\newblock URL \url{http://dx.doi.org/10.3390/jimaging8120320}.

\bibitem[Azizi et~al.(2021)Azizi, Mustafa, Ryan, Beaver, Freyberg, Deaton, Loh, Karthikesalingam, Kornblith, Chen, Natarajan, and Norouzi]{Azizi_2021_ICCV}
Shekoofeh Azizi, Basil Mustafa, Fiona Ryan, Zachary Beaver, Jan Freyberg, Jonathan Deaton, Aaron Loh, Alan Karthikesalingam, Simon Kornblith, Ting Chen, Vivek Natarajan, and Mohammad Norouzi.
\newblock Big self-supervised models advance medical image classification.
\newblock In \emph{Proceedings of the IEEE/CVF International Conference on Computer Vision (ICCV)}, pages 3478--3488, October 2021.

\bibitem[Bandi et~al.(2018)Bandi, Geessink, Manson, Van~Dijk, Balkenhol, Hermsen, Bejnordi, Lee, Paeng, Zhong, et~al.]{bandi2018detection}
Peter Bandi, Oscar Geessink, Quirine Manson, Marcory Van~Dijk, Maschenka Balkenhol, Meyke Hermsen, Babak~Ehteshami Bejnordi, Byungjae Lee, Kyunghyun Paeng, Aoxiao Zhong, et~al.
\newblock From detection of individual metastases to classification of lymph node status at the patient level: the camelyon17 challenge.
\newblock \emph{IEEE transactions on medical imaging}, 38\penalty0 (2):\penalty0 550--560, 2018.

\bibitem[Bejnordi et~al.(2017)Bejnordi, Veta, Van~Diest, Van~Ginneken, Karssemeijer, Litjens, Van Der~Laak, Hermsen, Manson, Balkenhol, et~al.]{bejnordi2017diagnostic}
Babak~Ehteshami Bejnordi, Mitko Veta, Paul~Johannes Van~Diest, Bram Van~Ginneken, Nico Karssemeijer, Geert Litjens, Jeroen~AWM Van Der~Laak, Meyke Hermsen, Quirine~F Manson, Maschenka Balkenhol, et~al.
\newblock Diagnostic assessment of deep learning algorithms for detection of lymph node metastases in women with breast cancer.
\newblock \emph{Jama}, 318\penalty0 (22):\penalty0 2199--2210, 2017.

\bibitem[Butoi et~al.(2023)Butoi, Ortiz, Ma, Sabuncu, Guttag, and Dalca]{butoi2023universeg}
Victor~Ion Butoi, Jose Javier~Gonzalez Ortiz, Tianyu Ma, Mert~R. Sabuncu, John Guttag, and Adrian~V. Dalca.
\newblock Universeg: Universal medical image segmentation, 2023.

\bibitem[Chaitanya et~al.(2020)Chaitanya, Erdil, Karani, and Konukoglu]{NEURIPS2020_949686ec}
Krishna Chaitanya, Ertunc Erdil, Neerav Karani, and Ender Konukoglu.
\newblock Contrastive learning of global and local features for medical image segmentation with limited annotations.
\newblock In H.~Larochelle, M.~Ranzato, R.~Hadsell, M.F. Balcan, and H.~Lin, editors, \emph{Advances in Neural Information Processing Systems}, volume~33, pages 12546--12558. Curran Associates, Inc., 2020.
\newblock URL \url{https://proceedings.neurips.cc/paper_files/paper/2020/file/949686ecef4ee20a62d16b4a2d7ccca3-Paper.pdf}.

\bibitem[Chen et~al.(2020{\natexlab{a}})Chen, Kornblith, Norouzi, and Hinton]{chen2020simple}
Ting Chen, Simon Kornblith, Mohammad Norouzi, and Geoffrey Hinton.
\newblock A simple framework for contrastive learning of visual representations.
\newblock In \emph{International conference on machine learning}, pages 1597--1607. PMLR, 2020{\natexlab{a}}.

\bibitem[Chen et~al.(2020{\natexlab{b}})Chen, Kornblith, Swersky, Norouzi, and Hinton]{chen2020big}
Ting Chen, Simon Kornblith, Kevin Swersky, Mohammad Norouzi, and Geoffrey~E Hinton.
\newblock Big self-supervised models are strong semi-supervised learners.
\newblock \emph{Advances in neural information processing systems}, 33:\penalty0 22243--22255, 2020{\natexlab{b}}.

\bibitem[Chen et~al.(2020{\natexlab{c}})Chen, Fan, Girshick, and He]{chen2020improved}
Xinlei Chen, Haoqi Fan, Ross Girshick, and Kaiming He.
\newblock Improved baselines with momentum contrastive learning.
\newblock \emph{arXiv preprint arXiv:2003.04297}, 2020{\natexlab{c}}.

\bibitem[Chlap et~al.(2021)Chlap, Min, Vandenberg, Dowling, Holloway, and Haworth]{chlap}
Phillip Chlap, Hang Min, Nym Vandenberg, Jason Dowling, Lois Holloway, and Annette Haworth.
\newblock A review of medical image data augmentation techniques for deep learning applications.
\newblock \emph{Journal of Medical Imaging and Radiation Oncology}, 65\penalty0 (5):\penalty0 545--563, 2021.
\newblock \doi{https://doi.org/10.1111/1754-9485.13261}.
\newblock URL \url{https://onlinelibrary.wiley.com/doi/abs/10.1111/1754-9485.13261}.

\bibitem[Deng et~al.(2009)Deng, Dong, Socher, Li, Li, and Fei-Fei]{deng2009imagenet}
Jia Deng, Wei Dong, Richard Socher, Li-Jia Li, Kai Li, and Li~Fei-Fei.
\newblock Imagenet: A large-scale hierarchical image database.
\newblock In \emph{2009 IEEE conference on computer vision and pattern recognition}, pages 248--255. Ieee, 2009.

\bibitem[Eaton-Rosen et~al.(2018)Eaton-Rosen, Bragman, Ourselin, and Cardoso]{eaton2018improving}
Zach Eaton-Rosen, Felix Bragman, Sebastien Ourselin, and M~Jorge Cardoso.
\newblock Improving data augmentation for medical image segmentation.
\newblock \emph{MIDL Abstract}, 2018.

\bibitem[Frid-Adar et~al.(2018{\natexlab{a}})Frid-Adar, Diamant, Klang, Amitai, Goldberger, and Greenspan]{FRIDADAR2018321}
Maayan Frid-Adar, Idit Diamant, Eyal Klang, Michal Amitai, Jacob Goldberger, and Hayit Greenspan.
\newblock Gan-based synthetic medical image augmentation for increased cnn performance in liver lesion classification.
\newblock \emph{Neurocomputing}, 321:\penalty0 321--331, 2018{\natexlab{a}}.
\newblock ISSN 0925-2312.
\newblock \doi{https://doi.org/10.1016/j.neucom.2018.09.013}.
\newblock URL \url{https://www.sciencedirect.com/science/article/pii/S0925231218310749}.

\bibitem[Frid-Adar et~al.(2018{\natexlab{b}})Frid-Adar, Klang, Amitai, Goldberger, and Greenspan]{8363576}
Maayan Frid-Adar, Eyal Klang, Michal Amitai, Jacob Goldberger, and Hayit Greenspan.
\newblock Synthetic data augmentation using gan for improved liver lesion classification.
\newblock In \emph{2018 IEEE 15th International Symposium on Biomedical Imaging (ISBI 2018)}, pages 289--293, 2018{\natexlab{b}}.
\newblock \doi{10.1109/ISBI.2018.8363576}.

\bibitem[Garcea et~al.(2023)Garcea, Serra, Lamberti, and Morra]{GARCEA2023106391}
Fabio Garcea, Alessio Serra, Fabrizio Lamberti, and Lia Morra.
\newblock Data augmentation for medical imaging: A systematic literature review.
\newblock \emph{Computers in Biology and Medicine}, 152:\penalty0 106391, 2023.
\newblock ISSN 0010-4825.
\newblock \doi{https://doi.org/10.1016/j.compbiomed.2022.106391}.
\newblock URL \url{https://www.sciencedirect.com/science/article/pii/S001048252201099X}.

\bibitem[Goodfellow et~al.(2020)Goodfellow, Pouget-Abadie, Mirza, Xu, Warde-Farley, Ozair, Courville, and Bengio]{goodfellow2020generative}
Ian Goodfellow, Jean Pouget-Abadie, Mehdi Mirza, Bing Xu, David Warde-Farley, Sherjil Ozair, Aaron Courville, and Yoshua Bengio.
\newblock Generative adversarial networks.
\newblock \emph{Communications of the ACM}, 63\penalty0 (11):\penalty0 139--144, 2020.

\bibitem[Gupta et~al.(2021)Gupta, Singh, Chang, Qu, Aggarwal, Arun, Vaswani, Raghavan, Agarwal, Gidwani, et~al.]{gupta2021addressing}
Sharut Gupta, Praveer Singh, Ken Chang, Liangqiong Qu, Mehak Aggarwal, Nishanth Arun, Ashwin Vaswani, Shruti Raghavan, Vibha Agarwal, Mishka Gidwani, et~al.
\newblock Addressing catastrophic forgetting for medical domain expansion.
\newblock \emph{arXiv preprint arXiv:2103.13511}, 2021.

\bibitem[He et~al.(2016)He, Zhang, Ren, and Sun]{he2016deep}
Kaiming He, Xiangyu Zhang, Shaoqing Ren, and Jian Sun.
\newblock Deep residual learning for image recognition.
\newblock In \emph{Proceedings of the IEEE conference on computer vision and pattern recognition}, pages 770--778, 2016.

\bibitem[He et~al.(2020)He, Fan, Wu, Xie, and Girshick]{he2020momentum}
Kaiming He, Haoqi Fan, Yuxin Wu, Saining Xie, and Ross Girshick.
\newblock Momentum contrast for unsupervised visual representation learning.
\newblock In \emph{Proceedings of the IEEE/CVF conference on computer vision and pattern recognition}, pages 9729--9738, 2020.

\bibitem[Huang et~al.(2017)Huang, Liu, Van Der~Maaten, and Weinberger]{huang2017densely}
Gao Huang, Zhuang Liu, Laurens Van Der~Maaten, and Kilian~Q Weinberger.
\newblock Densely connected convolutional networks.
\newblock In \emph{Proceedings of the IEEE conference on computer vision and pattern recognition}, pages 4700--4708, 2017.

\bibitem[Irvin et~al.(2019)Irvin, Rajpurkar, Ko, Yu, Ciurea-Ilcus, Chute, Marklund, Haghgoo, Ball, Shpanskaya, et~al.]{irvin2019chexpert}
Jeremy Irvin, Pranav Rajpurkar, Michael Ko, Yifan Yu, Silviana Ciurea-Ilcus, Chris Chute, Henrik Marklund, Behzad Haghgoo, Robyn Ball, Katie Shpanskaya, et~al.
\newblock Chexpert: A large chest radiograph dataset with uncertainty labels and expert comparison.
\newblock In \emph{Proceedings of the AAAI conference on artificial intelligence}, volume~33, pages 590--597, 2019.

\bibitem[Johnson et~al.(2019{\natexlab{a}})Johnson, Pollard, Greenbaum, Lungren, ying Deng, Peng, Lu, Mark, Berkowitz, and Horng]{johnson2019mimiccxrjpg}
Alistair E.~W. Johnson, Tom~J. Pollard, Nathaniel~R. Greenbaum, Matthew~P. Lungren, Chih ying Deng, Yifan Peng, Zhiyong Lu, Roger~G. Mark, Seth~J. Berkowitz, and Steven Horng.
\newblock Mimic-cxr-jpg, a large publicly available database of labeled chest radiographs, 2019{\natexlab{a}}.

\bibitem[Johnson et~al.(2019{\natexlab{b}})Johnson, Pollard, Berkowitz, Greenbaum, Lungren, Deng, Mark, and Horng]{johnson2019mimic}
Alistair~EW Johnson, Tom~J Pollard, Seth~J Berkowitz, Nathaniel~R Greenbaum, Matthew~P Lungren, Chih-ying Deng, Roger~G Mark, and Steven Horng.
\newblock Mimic-cxr, a de-identified publicly available database of chest radiographs with free-text reports.
\newblock \emph{Scientific data}, 6\penalty0 (1):\penalty0 317, 2019{\natexlab{b}}.

\bibitem[Ke et~al.(2021)Ke, Ellsworth, Banerjee, Ng, and Rajpurkar]{10.1145/3450439.3451867}
Alexander Ke, William Ellsworth, Oishi Banerjee, Andrew~Y. Ng, and Pranav Rajpurkar.
\newblock Chextransfer: Performance and parameter efficiency of imagenet models for chest x-ray interpretation.
\newblock In \emph{Proceedings of the Conference on Health, Inference, and Learning}, CHIL '21, page 116–124, New York, NY, USA, 2021. Association for Computing Machinery.
\newblock ISBN 9781450383592.
\newblock \doi{10.1145/3450439.3451867}.
\newblock URL \url{https://doi.org/10.1145/3450439.3451867}.

\bibitem[Krishnan et~al.(2022)Krishnan, Rajpurkar, and Topol]{krishnan2022self}
Rayan Krishnan, Pranav Rajpurkar, and Eric~J Topol.
\newblock Self-supervised learning in medicine and healthcare.
\newblock \emph{Nature Biomedical Engineering}, 6\penalty0 (12):\penalty0 1346--1352, 2022.

\bibitem[Madani et~al.(2018)Madani, Moradi, Karargyris, and Syeda-Mahmood]{10.1117/12.2293971}
Ali Madani, Mehdi Moradi, Alexandros Karargyris, and Tanveer Syeda-Mahmood.
\newblock {Chest x-ray generation and data augmentation for cardiovascular abnormality classification}.
\newblock In Elsa~D. Angelini and Bennett~A. Landman, editors, \emph{Medical Imaging 2018: Image Processing}, volume 10574, page 105741M. International Society for Optics and Photonics, SPIE, 2018.
\newblock \doi{10.1117/12.2293971}.
\newblock URL \url{https://doi.org/10.1117/12.2293971}.

\bibitem[Misra and Maaten(2020)]{misra2020self}
Ishan Misra and Laurens van~der Maaten.
\newblock Self-supervised learning of pretext-invariant representations.
\newblock In \emph{Proceedings of the IEEE/CVF conference on computer vision and pattern recognition}, pages 6707--6717, 2020.

\bibitem[Nguyen et~al.(2021)Nguyen, Pham, Nguyen, Nguyen, Huynh, Dao, and Vu]{nguyen2021vindr}
Hieu~T Nguyen, Hieu~H Pham, Nghia~T Nguyen, Ha~Q Nguyen, Thang~Q Huynh, Minh Dao, and Van Vu.
\newblock Vindr-spinexr: A deep learning framework for spinal lesions detection and classification from radiographs.
\newblock In \emph{Medical Image Computing and Computer Assisted Intervention--MICCAI 2021: 24th International Conference, Strasbourg, France, September 27--October 1, 2021, Proceedings, Part V 24}, pages 291--301. Springer, 2021.

\bibitem[Petersen et~al.(2010)Petersen, Aisen, Beckett, Donohue, Gamst, Harvey, Jack, Jagust, Shaw, Toga, et~al.]{petersen2010alzheimer}
Ronald~Carl Petersen, Paul~S Aisen, Laurel~A Beckett, Michael~C Donohue, Anthony~Collins Gamst, Danielle~J Harvey, Clifford~R Jack, William~J Jagust, Leslie~M Shaw, Arthur~W Toga, et~al.
\newblock Alzheimer's disease neuroimaging initiative (adni): clinical characterization.
\newblock \emph{Neurology}, 74\penalty0 (3):\penalty0 201--209, 2010.

\bibitem[Raghu et~al.(2017)Raghu, Gilmer, Yosinski, and Sohl-Dickstein]{raghu2017svcca}
Maithra Raghu, Justin Gilmer, Jason Yosinski, and Jascha Sohl-Dickstein.
\newblock Svcca: Singular vector canonical correlation analysis for deep learning dynamics and interpretability.
\newblock \emph{Advances in neural information processing systems}, 30, 2017.

\bibitem[Raghu et~al.(2019)Raghu, Zhang, Kleinberg, and Bengio]{NEURIPS2019_eb1e7832}
Maithra Raghu, Chiyuan Zhang, Jon Kleinberg, and Samy Bengio.
\newblock Transfusion: Understanding transfer learning for medical imaging.
\newblock In H.~Wallach, H.~Larochelle, A.~Beygelzimer, F.~d\textquotesingle Alch\'{e}-Buc, E.~Fox, and R.~Garnett, editors, \emph{Advances in Neural Information Processing Systems}, volume~32. Curran Associates, Inc., 2019.
\newblock URL \url{https://proceedings.neurips.cc/paper_files/paper/2019/file/eb1e78328c46506b46a4ac4a1e378b91-Paper.pdf}.

\bibitem[Rombach et~al.(2022)Rombach, Blattmann, Lorenz, Esser, and Ommer]{rombach2022high}
Robin Rombach, Andreas Blattmann, Dominik Lorenz, Patrick Esser, and Bj{\"o}rn Ommer.
\newblock High-resolution image synthesis with latent diffusion models.
\newblock In \emph{Proceedings of the IEEE/CVF conference on computer vision and pattern recognition}, pages 10684--10695, 2022.

\bibitem[Seyyed-Kalantari et~al.(2020)Seyyed-Kalantari, Liu, McDermott, Chen, and Ghassemi]{seyyed2020chexclusion}
Laleh Seyyed-Kalantari, Guanxiong Liu, Matthew McDermott, Irene~Y Chen, and Marzyeh Ghassemi.
\newblock Chexclusion: Fairness gaps in deep chest x-ray classifiers.
\newblock In \emph{BIOCOMPUTING 2021: proceedings of the Pacific symposium}, pages 232--243. World Scientific, 2020.

\bibitem[Sowrirajan et~al.(2021)Sowrirajan, Yang, Ng, and Rajpurkar]{sowrirajan2021mococxr}
Hari Sowrirajan, Jingbo Yang, Andrew~Y. Ng, and Pranav Rajpurkar.
\newblock Moco-cxr: Moco pretraining improves representation and transferability of chest x-ray models, 2021.

\bibitem[Stein et~al.(2019)Stein, Carol, Carr, Shih, Kalpathy-Cramer, Elliott, kalpathy, Prevedello, Kohli, Lungren, Culliton, Ball, and Halabi]{rsna-intracranial-hemorrhage-detection}
Anouk Stein, Wu~Carol Carol, Chris Carr, George Shih, Jayashree Kalpathy-Cramer, Julia Elliott, kalpathy, Luciano Prevedello, Marc Kohli, Matt Lungren, Phil Culliton, Robyn Ball, and Safwan Halabi.
\newblock Rsna intracranial hemorrhage detection, 2019.
\newblock URL \url{https://kaggle.com/competitions/rsna-intracranial-hemorrhage-detection}.

\bibitem[Szegedy et~al.(2015)Szegedy, Liu, Jia, Sermanet, Reed, Anguelov, Erhan, Vanhoucke, and Rabinovich]{szegedy2015going}
Christian Szegedy, Wei Liu, Yangqing Jia, Pierre Sermanet, Scott Reed, Dragomir Anguelov, Dumitru Erhan, Vincent Vanhoucke, and Andrew Rabinovich.
\newblock Going deeper with convolutions.
\newblock In \emph{Proceedings of the IEEE conference on computer vision and pattern recognition}, pages 1--9, 2015.

\bibitem[Thambawita et~al.(2022)Thambawita, Salehi, Sheshkal, Hicks, Hammer, Parasa, Lange, Halvorsen, and Riegler]{10.1371/journal.pone.0267976}
Vajira Thambawita, Pegah Salehi, Sajad~Amouei Sheshkal, Steven~A. Hicks, Hugo~L. Hammer, Sravanthi Parasa, Thomas~de Lange, Pål Halvorsen, and Michael~A. Riegler.
\newblock Singan-seg: Synthetic training data generation for medical image segmentation.
\newblock \emph{PLOS ONE}, 17\penalty0 (5):\penalty0 1--24, 05 2022.
\newblock \doi{10.1371/journal.pone.0267976}.
\newblock URL \url{https://doi.org/10.1371/journal.pone.0267976}.

\bibitem[Trabucco et~al.(2023)Trabucco, Doherty, Gurinas, and Salakhutdinov]{trabucco2023effective}
Brandon Trabucco, Kyle Doherty, Max Gurinas, and Ruslan Salakhutdinov.
\newblock Effective data augmentation with diffusion models.
\newblock \emph{arXiv preprint arXiv:2302.07944}, 2023.

\bibitem[Van~Essen et~al.(2013)Van~Essen, Smith, Barch, Behrens, Yacoub, Ugurbil, Consortium, et~al.]{van2013wu}
David~C Van~Essen, Stephen~M Smith, Deanna~M Barch, Timothy~EJ Behrens, Essa Yacoub, Kamil Ugurbil, Wu-Minn~HCP Consortium, et~al.
\newblock The wu-minn human connectome project: an overview.
\newblock \emph{Neuroimage}, 80:\penalty0 62--79, 2013.

\bibitem[Wang et~al.(2017)Wang, Peng, Lu, Lu, Bagheri, and Summers]{wang2017chestx}
Xiaosong Wang, Yifan Peng, Le~Lu, Zhiyong Lu, Mohammadhadi Bagheri, and Ronald~M Summers.
\newblock Chestx-ray8: Hospital-scale chest x-ray database and benchmarks on weakly-supervised classification and localization of common thorax diseases.
\newblock In \emph{Proceedings of the IEEE conference on computer vision and pattern recognition}, pages 2097--2106, 2017.

\bibitem[Wen et~al.(2021)Wen, Chen, Deng, and Zhou]{WEN2021103145}
Yang Wen, Leiting Chen, Yu~Deng, and Chuan Zhou.
\newblock Rethinking pre-training on medical imaging.
\newblock \emph{Journal of Visual Communication and Image Representation}, 78:\penalty0 103145, 2021.
\newblock ISSN 1047-3203.
\newblock \doi{https://doi.org/10.1016/j.jvcir.2021.103145}.
\newblock URL \url{https://www.sciencedirect.com/science/article/pii/S1047320321000894}.

\bibitem[Willemink et~al.(2020)Willemink, Koszek, Hardell, Wu, Fleischmann, Harvey, Folio, Summers, Rubin, and Lungren]{doi:10.1148/radiol.2020192224}
Martin~J. Willemink, Wojciech~A. Koszek, Cailin Hardell, Jie Wu, Dominik Fleischmann, Hugh Harvey, Les~R. Folio, Ronald~M. Summers, Daniel~L. Rubin, and Matthew~P. Lungren.
\newblock Preparing medical imaging data for machine learning.
\newblock \emph{Radiology}, 295\penalty0 (1):\penalty0 4--15, 2020.
\newblock \doi{10.1148/radiol.2020192224}.
\newblock URL \url{https://doi.org/10.1148/radiol.2020192224}.
\newblock PMID: 32068507.

\bibitem[Xie and Richmond(2018)]{Xie_2018_ECCV_Workshops}
Yiting Xie and David Richmond.
\newblock Pre-training on grayscale imagenet improves medical image classification.
\newblock In \emph{Proceedings of the European Conference on Computer Vision (ECCV) Workshops}, September 2018.

\bibitem[Zhang et~al.(2017)Zhang, Cisse, Dauphin, and Lopez-Paz]{zhang2017mixup}
Hongyi Zhang, Moustapha Cisse, Yann~N Dauphin, and David Lopez-Paz.
\newblock mixup: Beyond empirical risk minimization.
\newblock \emph{arXiv preprint arXiv:1710.09412}, 2017.

\bibitem[Zhou et~al.(2023)Zhou, Chen, and Lipton]{zhou2023evaluating}
Helen Zhou, Yuwen Chen, and Zachary Lipton.
\newblock Evaluating model performance in medical datasets over time.
\newblock In \emph{Conference on Health, Inference, and Learning}, pages 498--508. PMLR, 2023.

\end{thebibliography}

\appendix

\newpage
\onecolumn

\section{Hyperparameters}\label{app:training_details}
For linear finetuning, 100 epochs are enough for the model to converge, but for end-to-end finetuning, we try

25, 50, 100, and 200 epochs since it is easy to overfit.

Additional hyperparameters are shown in the table below:

\begin{table*}[ht]
\begin{adjustbox}{width=1\textwidth}
    \centering
    \begin{tabular}{ccccccc}
\toprule
Experiment & Dataset & Epochs & Batch size & Initial LR & Optimizer & LR Schedulers \\
\midrule
MoCo pretraining & VinDr-SpineXR & 500 & 64 & 0.002 & SGD (Weight Decay=$1e^{-4}$, momentum=$0.9$) & Cosine Annealing \\
MoCo pretraining & MIMIC-CXR & 100 & 256 & 0.002 & SGD (Weight Decay=$1e^{-4}$, momentum=$0.9$) & Cosine Annealing \\
Finetuning & VinDr-SpineXR & 25, 50, 100, 200 & 16 & 0.001 & SGD (Weight Decay=$1e^{-4}$, momentum=$0.9$) & Cosine Annealing \\
Finetuning & MIMIC-CXR & 25, 50, 100, 200 & 48 & 0.0005 & Adam & Cosine Annealing \\
\bottomrule
\end{tabular}
\end{adjustbox}
\caption{Training details for both MoCo pre-training and finetuning}
\label{tab:training_detail}
\end{table*}

\vspace{-2em}

\section{Impact of epochs on end-to-end finetuning performance}\label{app:end_to_end_vs_epoch}

\begin{figure}[ht]
    \centering
    \includegraphics[width=0.8\columnwidth]{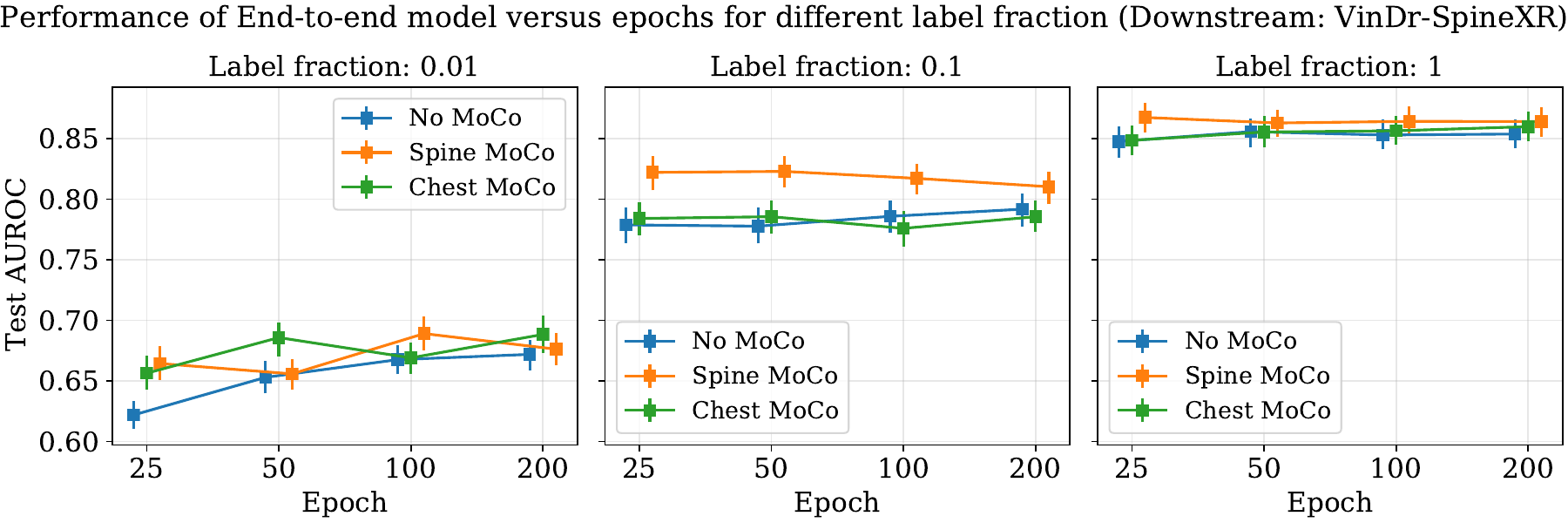}
    \caption{The test AUROC on VinDr-SpineXR of the models finetuned end-to-end vs. number of training epochs. The models are initialized by supervised pre-training on ImageNet, MoCo pre-training on VinDr-SpineXR (Spine) or MoCo pretraining on MIMIC-CXR (Chest).}
    \label{fig:vindr_vs_epoch}
\end{figure}

\begin{figure}[ht]
    \centering
    \includegraphics[width=0.95\columnwidth]{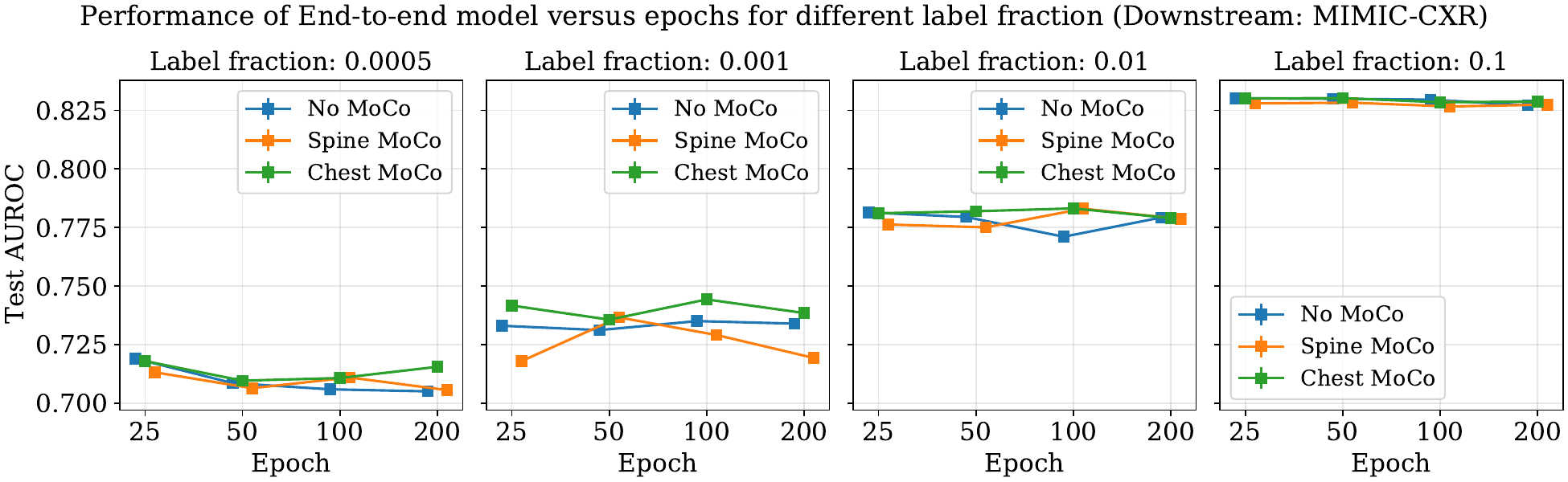}
    \caption{The test AUROC on MIMIC-CXR of the models finetuned end-to-end vs. number of training epochs. The models are initialized by supervised pre-training on ImageNet, MoCo pre-training on VinDr-SpineXR (Spine) or MoCo pretraining on MIMIC-CXR (Chest).}
    \label{fig:mimiccxr_vs_epoch}
\end{figure}

\newpage
\section{Test AUROC on VinDr-SpineXR and MIMIC-CXR}

\begin{table*}[ht]
\setlength{\tabcolsep}{4.5pt}
\small
    \centering
    \begin{adjustbox}{width=1\textwidth}

\begin{tabular}{ccccccc}
\toprule
 & \multicolumn{3}{c}{\textbf{MoCo + Linear Finetuning on Spine}} & \multicolumn{3}{c}{\textbf{MoCo + End-to-end Finetuning on Spine}} \\
Lab. \% (\#) &        No MoCo &          Spine MoCo &          Chest MoCo  &           No MoCo &             Spine MoCo &             Chest MoCo \\
\midrule
1\% (58) &  0.623 (0.610--0.634) &  \textbf{0.698 (0.683--0.714)} &  0.678 (0.665--0.694) &  0.668 (0.656--0.680) &  0.689 (0.675--0.703) &  0.669 (0.656--0.682) \\
10\% (587) &  0.758 (0.742--0.773) &  \textbf{0.818 (0.803--0.832)} &  0.782 (0.767--0.796) &  0.786 (0.772--0.799) &  0.817 (0.804--0.829) &  0.776 (0.761--0.790) \\
100\% (5872) &  0.799 (0.783--0.814) &  0.852 (0.839--0.867) &  0.821 (0.806--0.836) &  0.853 (0.841--0.866) &  \textbf{0.864 (0.852--0.877)} &  0.856 (0.845--0.868) \\
\bottomrule
\end{tabular}
\end{adjustbox}
\caption{The test AUROC on VinDr-SpineXR for models finetuned linearly and end-to-end on different proportions of labeled data. The results are presented as median (5\% quantile--95\% quantile).}
\label{tab:vindr_table}
\end{table*}

\begin{table*}[ht]
\setlength{\tabcolsep}{4.5pt}
 \centering
\small
\begin{adjustbox}{width=1\textwidth}
\begin{tabular}{ccccccc}
\toprule
 & \multicolumn{3}{c}{\textbf{MoCo + Linear Finetuning on Chest}} & \multicolumn{3}{c}{\textbf{MoCo + End-to-end Finetuning on Chest}} \\

Lab. \% (\#) &        No MoCo &          Spine MoCo &          Chest MoCo  &           No MoCo &             Spine MoCo &             Chest MoCo \\
\midrule
0.05\% (112) &  0.662 (0.661--0.664) &  0.679 (0.677--0.681) &  0.714 (0.713 - 0.716) &  0.706 (0.704 - 0.707) &  \textbf{0.711 (0.710 - 0.713)} &  \textbf{0.711 (0.709 - 0.712)} \\
0.1\% (225) &  0.699 (0.698--0.701) &  0.697 (0.695--0.699) &  0.739 (0.737 - 0.740) &  0.735 (0.734 - 0.737) &  0.729 (0.727 - 0.731) &  \textbf{0.744 (0.743 - 0.746)} \\
1\% (2257) &  0.752 (0.751--0.754) &  0.752 (0.751--0.754) &  \textbf{0.786 (0.784 - 0.787)} &  0.771 (0.770 - 0.772) &  0.783 (0.782 - 0.785) &  0.783 (0.782 - 0.785) \\
10\% (22572) &  0.777 (0.775--0.778) &  0.767 (0.765--0.769) &  0.812 (0.810 - 0.813) &  \textbf{0.830 (0.828 - 0.831)} &  0.827 (0.825 - 0.828) &  0.828 (0.827 - 0.830) \\
\bottomrule
\end{tabular}
\end{adjustbox}
\caption{The test AUROC on MIMIC-CXR for models finetuned linearly and end-to-end on different proportions of labeled data. The results are presented as median (5\% quantile--95\% quantile).}
\label{tab:mimic_table}
\end{table*}

\end{document}